\documentclass[letterpaper, 10 pt, conference]{ieeeconf}  

\usepackage{cite}
\usepackage{bm}
\usepackage{makecell}
\usepackage{amsmath} 
\usepackage{amssymb}  
\usepackage{etoolbox}
\usepackage{amsfonts}
\usepackage{booktabs}
\usepackage{ multicol} 

\usepackage{graphics} 
\usepackage{epsfig} 
\usepackage{mathptmx} 
\usepackage{times} 
\usepackage{adjustbox}

\usepackage{mathrsfs}
\usepackage{indentfirst}
\usepackage{multirow}
\usepackage{adjustbox}

\usepackage{color}      
\usepackage{pifont}     
\usepackage{algorithmic}
\usepackage{graphicx}
\usepackage{textcomp}
\usepackage{bbding}

\usepackage[table,dvipsnames,xcdraw]{xcolor} 

\usepackage{colortbl} 
\usepackage{diagbox} 
\usepackage{siunitx}

\usepackage{algorithm}
\usepackage{algorithmic}

\usepackage{subcaption}

\colorlet{colorFst}{Green!25}       
\colorlet{colorSnd}{SpringGreen!45} 
\colorlet{colorTrd}{Yellow!30}      
\colorlet{colorLow}{darkgray!30}    
\definecolor{R1}{HTML}{E97451}
\definecolor{R2}{HTML}{008080}
\definecolor{R3}{HTML}{0047AB}
\colorlet{cmt}{darkgray!80}    
\colorlet{supp}{darkgray!50}    

\definecolor{cvprblue}{rgb}{0.21,0.49,0.74}
\definecolor{mimicorange}{rgb}{0.84,0.43,0.16}
\definecolor{mimicpurple}{RGB}{140,60,120}

\usepackage[colorlinks=true,linkcolor=blue,citecolor=green,urlcolor=blue,]{hyperref}
\definecolor{bgcolor}{RGB}{140, 60, 120}
\definecolor{mimicolive}{RGB}{95,120,45}

\definecolor{mimicgold}{RGB}{180,140,60}

\newcommand{\compactdisplaymathbegin}{\ifhmode\par\fi\begingroup\scriptsize}
\BeforeBeginEnvironment{equation}{\compactdisplaymathbegin}
\AfterEndEnvironment{equation}{\endgroup}
\BeforeBeginEnvironment{equation*}{\compactdisplaymathbegin}
\AfterEndEnvironment{equation*}{\endgroup}
\BeforeBeginEnvironment{align}{\compactdisplaymathbegin}
\AfterEndEnvironment{align}{\endgroup}
\BeforeBeginEnvironment{align*}{\compactdisplaymathbegin}
\AfterEndEnvironment{align*}{\endgroup}

\newcommand{\fs}{\cellcolor{colorFst}}   
\newcommand{\nd}{\cellcolor{colorSnd}}      
\newcommand{\rd}{\cellcolor{colorTrd}}      

\IEEEoverridecommandlockouts                              

\title{
  \textcolor{mimicpurple}{Dream}\textcolor{cvprblue}{Mimic}:
  Learning Visuomotor Whole-Body Loco-Manipulation\\
  via World Model
}

\IEEEaftertitletext{%
\vspace{-1\baselineskip}
\begin{center}
\textbf{Project Page: \url{https://dreammimic.github.io/}}
\end{center}
\vspace{0.1\baselineskip}
}

\author{Jie Yin$^{1,\dagger}$*, Xingyu Lai$^{2,\dagger}$
\thanks{$^{1}$Independent, $^{2}$Tsinghua University.$^{\dagger}$equal contribution.}%
}

\begin{document}

\maketitle
\thispagestyle{empty}
\pagestyle{empty}

\begin{abstract}

Vision-based whole-body loco-manipulation on humanoid robots is challenging due to partial observability, contact-rich dynamics, and the difficulty of learning long-horizon behaviors from high-dimensional visual inputs. We present \href{https://github.com/DreamMimic/DreamMimic}{DreamMimic}, a framework that distills privileged teacher policies into vision-based humanoid controllers via world-model-assisted distillation. Instead of using a Dreamer-style RSSM for planning, we repurpose it to learn predictive latent dynamics that serve as both a representation space and an action-conditioned multi-step supervision signal, while exposing compact predictive features to the student policy to reduce long-term drift.
Beyond standard reconstruction objectives for proprioceptive and visual observations, we add auxiliary prediction heads for privileged state, contact, object state, and reward estimation. These heads provide additional supervision related to agent--object interaction and task progress, encouraging the latent representation to retain signals that are useful for contact-rich loco-manipulation.
We further introduce Performance-Conditioned Guidance (PCG), a reward-driven adaptive distillation schedule that computes performance scores for both teacher and student to dynamically balance guidance and exploration. PCG prevents both premature teacher annealing and excessive teacher interference in challenging visual settings.
Experiments on OMOMO and BEHAVE show improved tracking-based loco-manipulation performance over strong vision-based baselines, without exposing online privileged interaction states to the student at deployment. Qualitative simulations further examine morphology and simulator changes. These results suggest that world models can provide a useful mechanism for stabilizing visual policy distillation in contact-rich humanoid behaviors.

\end{abstract}

\begin{figure}[t]
    \centering
    \includegraphics[width=0.9\columnwidth]{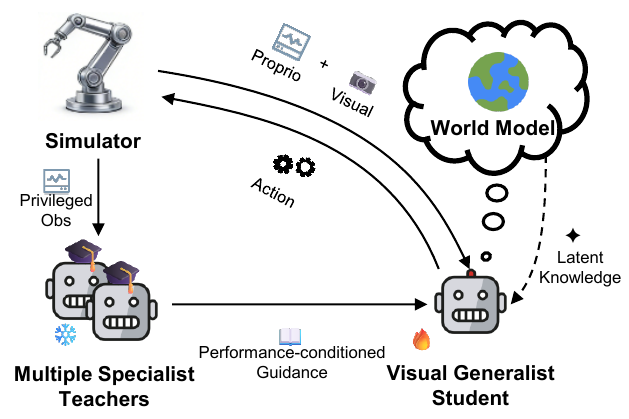}
    \caption{\textbf{Overview of the DreamMimic framework:}
    Multiple specialist teachers trained with privileged simulation observations are consolidated into a unified privileged teacher, which guides a vision-based student via \emph{Performance-Conditioned Guidance} (PCG). PCG adaptively balances teacher supervision and student exploration according to their relative performance.
    The student operates on non-privileged proprioception, a compact goal condition, and world-model features inferred from depth and segmentation. The goal condition jointly describes the target object pose and short-horizon robot trajectory cues, while the world model supplies predictive interaction features and action-conditioned multi-step latent supervision. This design enables the policy to exploit both task-level goals and inferred interaction cues while reducing long-horizon drift.
    }

    \label{pipeline}
    \vspace{-5mm}
\end{figure}

\begin{figure*}[t]
    \centering
    \includegraphics[width=1.75\columnwidth]{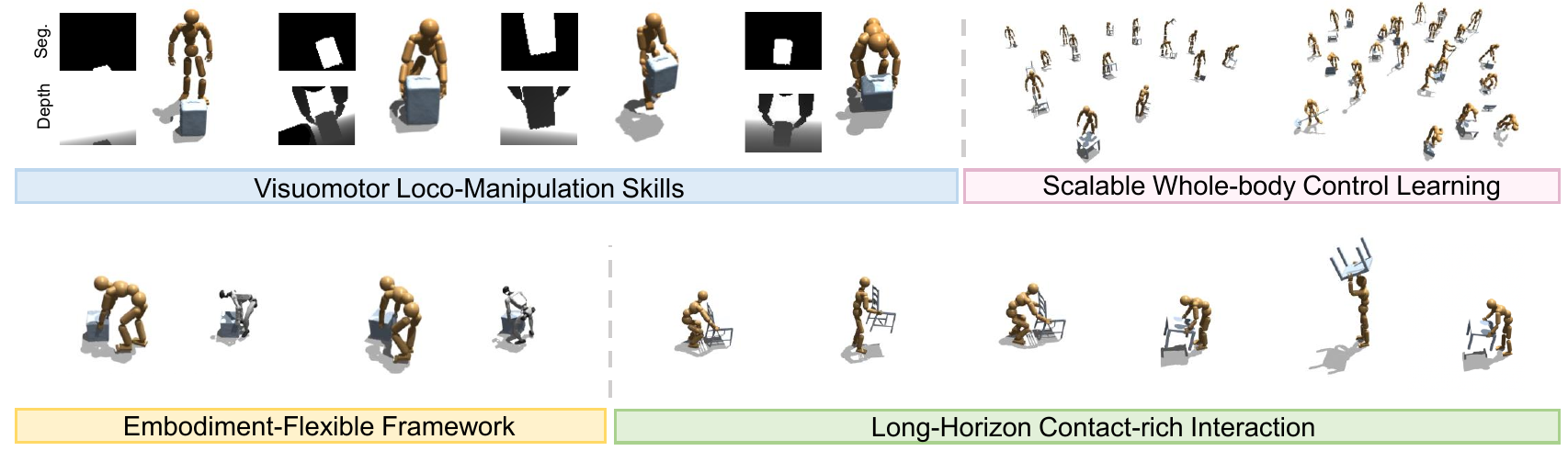}
    \caption{
        \textbf{DreamMimic supports vision-based humanoid loco-manipulation through world-model-assisted policy distillation.}
        Distilling privileged specialist teachers into a visual student enables whole-body interaction under partial observability (top left) and diverse behaviors (top right).
        The same framework transfers across morphologies, including SMPL-X and Unitree G1 (bottom left), and sustains long-horizon contact-rich loco-manipulation (bottom right).
        }
    \label{teaser}
    \vspace{-2mm}
\end{figure*}

\section{INTRODUCTION}
Humanoid robots promise transformative impact in factory automation and daily-life assistance, where they must perform whole-body loco-manipulation under visual feedback. Despite recent progress, learning robust visuomotor humanoid policies for such tasks remains difficult due to discontinuous contact dynamics, high-dimensional action spaces, and the partial observability of visual observations.

Prior work has advanced humanoid motion transfer and whole-body control, yet key limitations still remain. Many approaches focus on locomotion without complex object interactions\cite{videomimic}, while others rely on privileged information \cite{pan2025tokenhsi,xu2025intermimic} or pre-computed state estimators \cite{rouxel2022multicontact} rather than learning directly from vision. In whole-body loco-manipulation, such privileged signals, including ground-truth object pose, interaction geometry, and object contact, are especially important because they disambiguate contact timing and agent--object coupling; however, they are unavailable to a deployable visuomotor controller. Existing visuomotor \cite{zhao2025resmimic} frameworks often lack mechanisms to capture temporally structured, predictive representations, making them vulnerable to compounding errors in long-horizon contact-rich tasks.

Recent studies have shown the potential of world models for visual control in domains such as legged locomotion\cite{lai2024worldmodelbasedperceptionvisual}, autonomous driving\cite{jiang2025wpt}, and arm control\cite{yamada2024twist}. While these studies demonstrate the value of predictive latent dynamics for visual behaviors, they are evaluated on settings with lower‑dimensional action spaces or limited contact interactions. In this work, we focus on end-to-end visual distillation for high-degree-of-freedom humanoid loco-manipulation, where rich contact dynamics and temporal coordination make predictive representations and multi-step supervision particularly critical.

To bridge these gaps, we present \textbf{DreamMimic}, a world-model-assisted framework for stabilizing visuomotor policy distillation in humanoid loco-manipulation. Instead of using the world model for planning~\cite{hafner2019dream,hafner2020mastering,hafner2025mastering}, we repurpose a Dreamer-style RSSM to learn predictive state representations and provide multi-step supervision, mitigating compounding errors under partial observability. We further augment the world model with interaction-aware auxiliary objectives that estimate task-relevant interaction cues from student-side observations, rather than exposing online ground-truth contact or interaction states to the student. The pipeline of DreamMimic is overviewed in Figure~\ref{pipeline}. We highlight our contributions as follows:

\begin{itemize}

\item 
\textbf{DreamMimic: a world-model-assisted framework for stable visual policy distillation}. 
DreamMimic mitigates compounding errors under partial observability by leveraging a Dreamer-style RSSM to learn predictive state representations and enforce multi-step action-conditioned latent alignment, enabling temporally coherent behaviors for contact-rich humanoid loco-manipulation.

\item
\textbf{Structured supervision for stable DAgger+RL distillation}. 
We use auxiliary prediction heads to provide learning signals for reward, contact, and object dynamics, which help shape predictive representations for interaction-rich behaviors. Building on this supervision, \emph{Performance-Conditioned Guidance} (PCG) adaptively modulates teacher involvement based on relative teacher--student performance, stabilizing policy optimization while avoiding premature guidance decay.

\item 
\textbf{Comprehensive simulation evaluation across datasets, humanoids, and simulators}. Quantitative experiments on OMOMO and BEHAVE report consistent improvements in loco-manipulation performance, while qualitative simulations on the Unitree G1 and in Isaac Sim provide preliminary evidence on morphology and simulator changes (Fig.~\ref{teaser}). 

\end{itemize}

\section{Related Work}
\vspace{-1mm}

\subsection{Physics‑based Character Control}
\vspace{-1mm}

Physics-based imitation and reinforcement learning have enabled remarkable progress in whole-body character control within simulation, achieving high-fidelity motion reproduction~\cite{peng2018deepmimic,peng2021amp} and increasingly complex behaviors such as path following~\cite{rempe2023trace}, motion tracking~\cite{luo2023perpetual}, and highly dynamic skills including sports~\cite{yuan2023learning,wang2024skillmimic,xu2025learning}. Despite these successes, transferring such capabilities to real robots remains challenging, largely due to the reliance on privileged observations (e.g., contact states, dense task-state information, or object poses) that are unavailable in real-world deployment.
Building upon a recent teacher–student framework that scales imitation from large, imperfect datasets~\cite{xu2025intermimic}, we take a step further by distilling the privileged generalist into a vision-based student that operates on onboard sensing. Unlike prior approaches that emphasize locomotion-only skills, DreamMimic targets visuomotor distillation while preserving long-horizon policy fidelity, with the goal of enabling perception-driven whole-body behaviors for contact-rich loco-manipulation.

\vspace{-1mm}
\subsection{Visual Policy Learning for Robotic Control}
\vspace{-1mm}

Learning visuomotor policies directly with reinforcement learning is notoriously unstable due to high-dimensional observations and delayed credit assignment\cite{LevineFDA15}. Model-based RL with learned latent dynamics (e.g., Dreamer~\cite{hafner2019dream,hafner2020mastering}) can improve sample efficiency for learning from pixels by learning predictive representations and training policies through imagined rollouts. However, single-stage learning from vision remains challenging for contact-rich humanoid loco-manipulation, where partial observability, discontinuous contacts, and large action spaces exacerbate exploration and temporal dependencies. Some recent works utilize teacher–student paradigms to distill skills into vision-based policies for manipulation\cite{jiang2024transic} and quadruped loco-manipulation \cite{liu2024visual}, and student-aware teacher optimization has been explored to reduce teacher–student observability mismatch~\cite{messikommer2025studentinformed}. Adaptive combinations of imitation and reinforcement learning have also been studied through performance-based modulation~\cite{leiva2024combining}. In this work, we stabilize visual policy distillation by leveraging a predictive world model that provides temporally coherent supervision, mitigating distribution shift and multi-step error accumulation. We further combine DAgger \cite{2010dagger} with auxiliary PPO regularization, PCG, and an InterMimic-style reference-buffer curriculum to support stable acquisition of vision-based whole-body control skills (Fig.~\ref{teaser}).
Unlike iteration-based teacher annealing~\cite{xu2025intermimic}, PCG adapts the teacher-driven rollout ratio from reward-based relative performance while keeping the imitation coefficient fixed, following prior work on performance-conditioned IL--RL balancing~\cite{leiva2024combining} but targeting teacher--student rollout allocation under partial observability.

\vspace{-2mm}
\subsection{World Models for Robots}
\vspace{-1mm}

World models learn latent dynamics to summarize observation histories and predict future states for control and planning, improving data efficiency in high-dimensional settings. The Dreamer family learns a compact recurrent dynamics model and optimizes policies using imagined trajectories, enabling efficient learning from pixel observations~\cite{hafner2019dream,hafner2020mastering,hafner2025mastering}. DayDreamer~\cite{wu2023daydreamer} extends this paradigm to real robots, while recent work explores diffusion-based world models for operating policies entirely in imagined environments~\cite{jiang2025world4rl,li2025robotic}. Object-centric world models further factor observations into entity-level dynamics for interaction reasoning~\cite{nishimoto2026objectcentric}, and recent humanoid interaction work explicitly predicts object dynamics for agile control~\cite{li2026haic}.
However, experiments reveal that Dreamer-style agents struggle on high-dimensional visual humanoid tasks in HumanoidBench~\cite{sferrazza2024humanoidbench}. Prior studies mainly focus on planning or direct policy optimization, and rarely examine how world models can stabilize visual policy distillation. In contrast, DreamMimic leverages a recurrent world model as both a representation backbone and a multi-step supervision signal. Predictive latent dynamics regularize teacher–student alignment under partial observability, helping distill long-horizon contact behaviors (Figure~\ref{teaser}) into vision-based policies while complementing privileged-teacher distillation frameworks~\cite{xu2025intermimic}.

\section{Method}

We study contact-rich humanoid loco-manipulation, in which a robot coordinates whole-body motion with object interaction under partial observability. Given sequences of 153-DoF SMPL-X~\cite{pavlakos2019expressive} motion and object trajectories, our goal is to learn a visuomotor policy that reproduces physically consistent interactions without simulator-only privileged information at test time.
\vspace{-1mm}
\subsection{Problem Formulation}
\vspace{-1mm}

We formulate this problem as a partially observable Markov decision process (POMDP), where the underlying Markovian environment state is not directly accessible from onboard sensing. Specifically, let \(\boldsymbol{x}_t\) denote the full simulator state at time \(t\), \(\boldsymbol{a}_t\) the robot action, and \(\boldsymbol{r}_t\) the task reward. The deployed student observes visual and proprioceptive inputs and receives a compact goal condition, but it must infer interaction-relevant variables (e.g., contacts and object dynamics) from observation histories. We therefore seek a deployable goal-conditioned visuomotor policy that maps onboard sensing and task commands to actions. At test time, the student conditions on non-privileged proprioception, a compact goal condition containing target object pose and robot trajectory cues, and world-model features inferred from visual--proprioceptive histories, rather than on online privileged simulator states such as contact labels or interaction graphs. 

\vspace{-1mm}
\subsection{Privileged Teacher Policy}
\vspace{-1mm}

The teacher policy has access to full simulator state, including ground-truth object pose, interaction graph, and object contact signals\cite{wang2024skillmimic}, enabling stable optimization of complex contact behaviors. These signals provide direct supervision for when and how the body should couple with the object, which is critical for coordinated loco-manipulation but unavailable at deployment. The teacher is trained via reinforcement learning to track the reference motion while maintaining physically plausible dynamics.

The reward integrates whole-body tracking, end-effector alignment, object consistency, and contact-aware terms \cite{wang2024skillmimic,xu2025intermimic}, encouraging coordinated loco-manipulation rather than isolated skills. For large-scale Human-Object Interaction (HOI) datasets, we follow InterMimic’s specialist-to-generalist recipe~\cite{xu2025intermimic} to construct multiple privileged teachers across diverse interaction scenarios, and use them as the supervision source for the vision-based student.

\vspace{-2mm}
\subsection{Visuomotor Student Distillation}
\vspace{-1mm}

The student policy network does not directly encode raw images at training or test time. Instead, a world model encodes onboard depth and segmentation into predictive latent features, which are passed to the policy together with non-privileged proprioception and a compact goal condition.

Let \(\boldsymbol{p}_t\) denote the student-side proprioceptive observation after removing simulator-only object state, interaction graph, and object-contact labels. We denote the task command by \(\boldsymbol{g}_t\), which compactly combines target object pose and short-horizon robot trajectory cues, including root and key-body targets. This goal condition specifies what interaction should be tracked without exposing online simulator interaction states. The policy condition is then
\begin{align}
\boldsymbol{c}_t &= \left[\boldsymbol{p}_t,\boldsymbol{g}_t,\boldsymbol{\phi}^{\text{wm}}_t\right], \\
\boldsymbol{a}_t &\sim \pi_\theta(\cdot \mid \boldsymbol{c}_t),
\end{align}
where the world-model feature is
\begin{equation}
\boldsymbol{\phi}^{\text{wm}}_t =
\left[\boldsymbol{h}_t,\hat r_t,\hat{\boldsymbol{x}}^{\text{priv}}_t,\hat{\boldsymbol{c}}^{\text{contact}}_t,\hat{\boldsymbol{x}}^{\text{obj}}_t\right].
\end{equation}
Here \(\boldsymbol{h}_t\) is the deterministic RSSM state inferred from the visual--proprioceptive history and previous action, \(\hat r_t\) is the reward prediction, and the remaining terms are auxiliary predictions for privileged state, contact, and object state. These predicted quantities provide compact estimates of interaction-relevant cues; the student does not receive online ground-truth object state, interaction geometry, or contact labels as policy inputs. In our policy network, the proprioceptive-goal feature, deterministic world-model feature, and auxiliary prediction feature are projected as separate tokens and fused by a lightweight Transformer before the actor--critic MLP.
We perform action distillation using supervised imitation as the primary learning signal and use PPO regularization as an auxiliary stabilizer on student-driven rollouts.


\textbf{Performance-Conditioned Guidance (PCG).} 
PCG balances supervision and exploration by allocating environments to either \emph{teacher-driven} or \emph{student-driven} rollouts and adapting this allocation based on performance.
At the beginning of each episode, we sample an environment role indicator \(m\in\{0,1\}\), where \(m=1\) denotes that the teacher executes actions for that environment. Let \(\rho\in[0,1]\) denote the fraction of teacher-driven environments. During training, we maintain exponential moving averages of shaped rewards for teacher-driven and student-driven environments, denoted \(\hat r_T\) and \(\hat r_S\), and compute a relative performance score \(\pi=\hat r_S/(\hat r_T+\epsilon)\).
As \(\pi\) approaches a target performance ratio, we decay the teacher-environment ratio \(\rho\) from \(\rho_{\max}\) to \(\rho_{\min}\) using an EMA-smoothed update, while keeping the imitation supervision coefficient fixed throughout training. This prevents guidance from vanishing prematurely when visual policies are still under-trained, while avoiding excessive teacher interference once the student becomes competent under its own state distribution.


\textbf{Reference-buffer curriculum.} 
Student training uses an InterMimic-style physical reference-buffer curriculum. During training, failed student-driven rollouts are used to update a small set of replayed reference states around difficult time windows. Future rollouts can then revisit these difficult interaction states, while teacher-driven environments are excluded from this update so that the curriculum reflects the student's own state distribution. 
\vspace{-2mm}
\subsection{World Model for Predictive Representation}
\vspace{-1mm}

\begin{figure*}[ht]
    \centering
    \includegraphics[width=1.6\columnwidth]{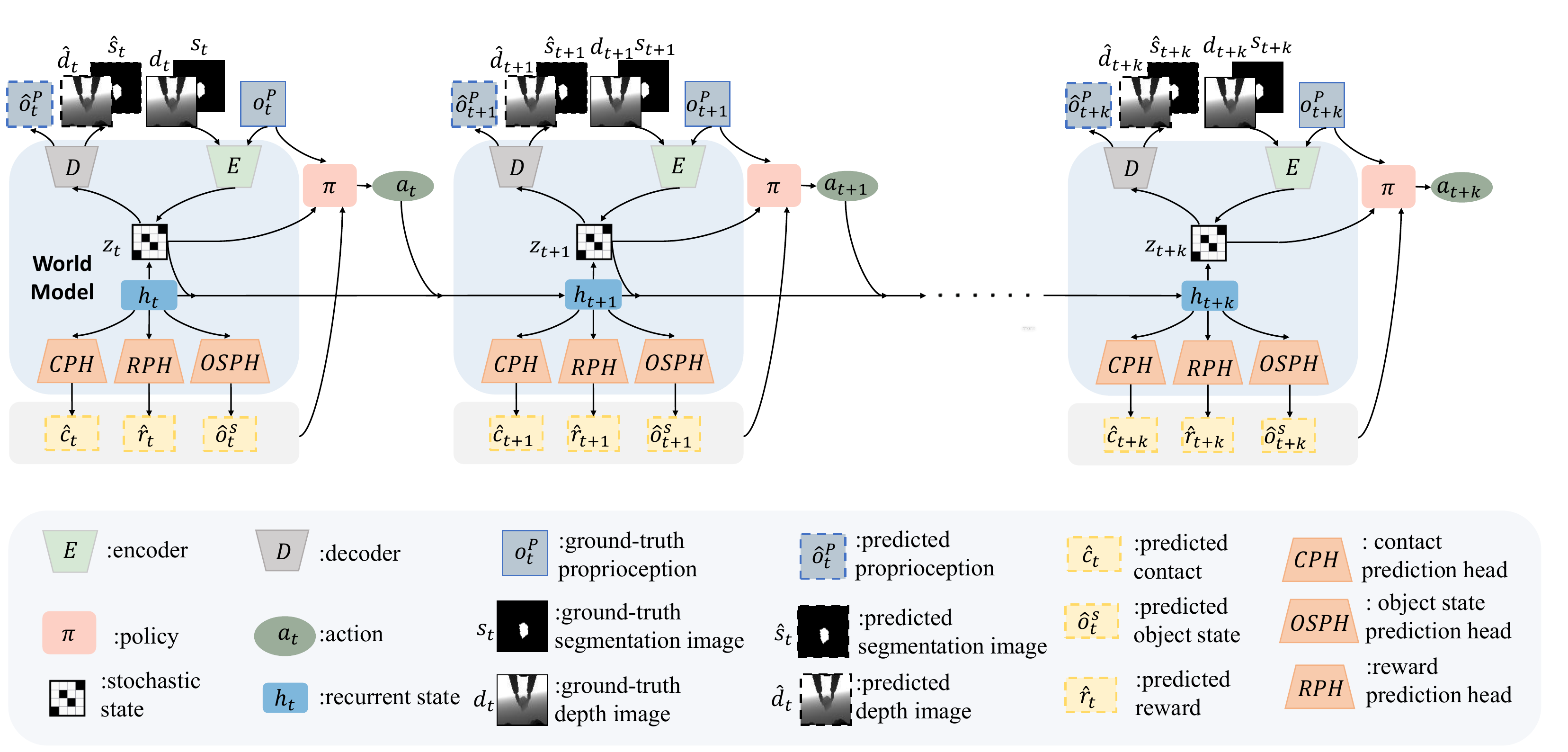}
    \caption{\textbf{Overview of the world model framework.} A Dreamer-style world model jointly learns predictive latent dynamics from visual and proprioceptive observations and provides action-conditioned multi-step latent targets to stabilize policy learning.} 
    \label{world_model}
    \vspace{-4mm}
\end{figure*}

Single-step imitation is insufficient in contact-rich settings because perceptual uncertainty can cause small errors that compound over time. To address this limitation, we learn a recurrent state-space world model based on the RSSM architecture, whose role is not planning but predictive representation learning for distillation. The structure of our world model is illustrated in Figure~\ref{world_model}.

\textbf{Notation.} At timestep \(t\), \(\boldsymbol{o}_t=(\boldsymbol{o}^{\text{vis}}_t,\boldsymbol{o}^{\text{prop}}_t)\) denotes the student observation (visual input and proprioception), and \(\boldsymbol{a}_{t-1}\) denotes the previous action.

\textbf{Encoder.} The world model encoder \(E_\xi\) maps the current observation \(\boldsymbol{o}_t\) and previous action \(\boldsymbol{a}_{t-1}\) to an embedding \(\boldsymbol{e}_t\):
\begin{equation}
\boldsymbol{e}_t = E_\xi(\boldsymbol{o}_t, \boldsymbol{a}_{t-1}),
\end{equation}
where visual inputs are processed by a lightweight convolutional encoder and proprioceptive inputs by an MLP. Temporal context is accumulated through the recurrent RSSM state rather than by stacking a long explicit observation history in the policy.

\textbf{RSSM latent dynamics.} The RSSM maintains a latent state \(\boldsymbol{z}_t = (\boldsymbol{h}_t, \boldsymbol{s}_t)\) with a deterministic recurrent state \(\boldsymbol{h}_t\) and a stochastic state \(\boldsymbol{s}_t\). The deterministic transition \(f_\xi\), implemented with a GRU, updates \(\boldsymbol{h}_t\) from the previous latent state and action, while the posterior \(q_\xi\) infers \(\boldsymbol{s}_t\) conditioned on \(\boldsymbol{h}_t\) and the observation embedding \(\boldsymbol{e}_t\):
\begin{align}
\boldsymbol{h}_t &= f_\xi(\boldsymbol{h}_{t-1}, \boldsymbol{s}_{t-1}, \boldsymbol{a}_{t-1}), \\
\boldsymbol{s}_t &\sim q_\xi(\boldsymbol{s}_t \mid \boldsymbol{h}_t, \boldsymbol{e}_t).
\end{align}

\textbf{Prior prediction.} To enable open-loop prediction without the current observation, the RSSM also defines a prior \(p_\xi\) over the stochastic state by conditioning only on the deterministic transition:
\begin{align}
\boldsymbol{h}_t^{\text{prior}} &= f_\xi(\boldsymbol{h}_{t-1}, \boldsymbol{s}_{t-1}, \boldsymbol{a}_{t-1}), \\
\boldsymbol{s}_t^{\text{prior}} &\sim p_\xi(\boldsymbol{s}_t \mid \boldsymbol{h}_t^{\text{prior}}).
\end{align}

\textbf{Decoding and auxiliary objectives.} From the latent state, we attach a reconstruction head (RecH) implemented as a decoder \(D_\xi\), together with auxiliary predictors for reward, privileged-state, contact, and object-state prediction. RecH reconstructs both visual and proprioceptive observations with image MSE and proprioceptive symlog-MSE losses, preserving sensory fidelity in the latent space during world-model training. However, RecH outputs are not fed to the policy; they serve only as representation-learning targets. The policy instead consumes the deterministic latent \(\boldsymbol{h}_t\) and the auxiliary predictions. The reward predictor estimates the instantaneous task reward. The privileged-state predictor estimates the teacher-side interaction target, formed by concatenating the 13-D object state, full-body interaction graph, and target contact label. The contact predictor estimates the target contact label, while the object-state predictor estimates the 13-D target object state. These supervision signals are extracted from simulator and task annotations during training; at test time, the student receives only their world-model predictions. Overall, \(\boldsymbol{z}_t\) summarizes visuoproprioceptive history and supports inference of unobservable interaction variables, providing temporally structured features for long-horizon policy learning.
\vspace{-2mm}
\subsection{Multi-step Latent Distillation}
\vspace{-1mm}

Standard distillation aligns actions only at the current timestep, leaving future state evolution unconstrained. We therefore add an action-conditioned latent consistency loss that compares how the world model predicts the consequences of student and teacher actions.

Starting from timestep $t$, we record the posterior latent state $\boldsymbol{z}_t=(\boldsymbol{h}_t,\boldsymbol{s}_t)$ inferred from the current student-side observation. We then branch two imagined rollouts from this same latent state through the RSSM prior while keeping the world-model parameters frozen. One branch is conditioned on the student policy mean action \(\boldsymbol{\mu}^{(S)}_t\), and the other is conditioned on the teacher policy mean action \(\boldsymbol{\mu}^{(T)}_t\). We repeat this action-conditioned prior rollout for \(H\) steps (we use \(H{=}3\) in all main experiments), without consuming additional observations. Let \(\boldsymbol{z}^{(S)}_{t+k}\) and \(\boldsymbol{z}^{(T)}_{t+k}\) denote the latent states induced by the student and teacher actions after \(k\) imagined steps. We minimize

\begin{equation}
\mathcal{L}_{\text{latent}} =
\sum_{k=1}^{H}
\left[ \| \boldsymbol{h}^{(S)}_{t+k} - \boldsymbol{h}^{(T)}_{t+k} \|_2^2 + \lambda_{\text{stoch}} \| \boldsymbol{s}^{(S)}_{t+k} - \boldsymbol{s}^{(T)}_{t+k} \|_2^2 \right],
\end{equation}
where $\lambda_{\text{stoch}}$ balances deterministic and stochastic matching. Matching the deterministic component $\boldsymbol{h}$ encourages alignment in the summary of history, while the stochastic-state matching penalizes divergence in the discrete RSSM state used by the policy feature. This loss does not require separate teacher-side visual observations; it uses the learned dynamics as a local, action-conditioned consistency metric between the student and teacher policy means.
\vspace{-2mm}
\subsection{Training Objective}
\vspace{-1mm}

The world model is trained on student experience to capture action-conditioned dynamics in a compact latent space. Its objective combines KL-regularized representation learning with auxiliary predictions that anchor the latent state to task-relevant signals:

\begin{equation}
\mathcal{L}_{\text{wm}} =
\mathcal{L}_{\text{KL}} +
\mathcal{L}_{\text{rec}} +
\mathcal{L}_{\text{rew}} +
\mathcal{L}_{\text{priv}} +
\mathcal{L}_{\text{con}} +
\mathcal{L}_{\text{obj}},
\end{equation}
where $\mathcal{L}_{\text{KL}} = \alpha_{\text{dyn}} \mathcal{L}_{\text{dyn}} + \alpha_{\text{rep}} \mathcal{L}_{\text{rep}}$ balances dynamics and representation learning. The KL loss includes a free-bits mechanism to prevent posterior collapse. The reconstruction loss $\mathcal{L}_{\text{rec}}$ reconstructs both visual and proprioceptive observations through RecH. The remaining terms supervise the instantaneous reward, the teacher-side privileged vector, the target-contact signal, and the target object state, respectively.

The overall optimization separates world-model representation learning from policy distillation. The student policy loss combines mean-action matching $\mathcal{L}_{\text{action}} = \| \boldsymbol{\mu}_t^{(S)} - \boldsymbol{\mu}_t^{(T)} \|_2^2$ with multi-step latent supervision $\mathcal{L}_{\text{latent}}$ and PPO regularization:
\begin{equation}
\mathcal{L}_{\text{total}} = c\, w_{\text{action}} \mathcal{L}_{\text{action}} + w_{\text{wm}} \mathcal{L}_{\text{latent}} + w_{\text{ppo}} \mathcal{L}_{\text{ppo}},
\end{equation}
where \(c\) denotes the imitation supervision coefficient. In our implementation, the optimization is imitation-dominant: supervised action matching remains the main policy-learning signal, while PPO is introduced after a warm-up period as a lower-weight regularizer for student-driven rollouts. Under PCG, teacher guidance is governed primarily by the teacher-driven environment ratio \(\rho\), which is adapted online from reward-based relative performance as described above. We keep \(\mathcal{L}_{\text{wm}}\) and \(\mathcal{L}_{\text{latent}}\) separate because they serve different roles: \(\mathcal{L}_{\text{wm}}\) trains the predictive model from observed student sequences, whereas \(\mathcal{L}_{\text{latent}}\) uses the frozen learned dynamics as an action-conditioned supervisory metric between student and teacher policy means. The world model is updated periodically from replayed rollout sequences, providing a stable predictive space while avoiding a single coupled objective in which the policy and representation targets drift simultaneously.

\section{Experiments}

\noindent\textbf{Datasets.} We use large-scale Human-Object Interaction (HOI) datasets processed following the pipeline of InterAct~\cite{xu2025interact}, which generates physically grounded reference trajectories by retargeting human motions to objects while enforcing geometric consistency and contact feasibility. Specifically, we employ OMOMO~\cite{li2023object}, which provides full-body human motions paired with object trajectories and contact information suitable for learning coordinated manipulation. From OMOMO, we select five objects with diverse shapes and scales: large table, wooden chair, plastic box, small box and suitcase. We also use BEHAVE~\cite{bhatnagar22behave}, a dataset featuring natural full-body human–object interactions with reliable body–object alignment. To emphasize sustained loco-manipulation, we focus on long-horizon sequences whose average duration exceeds 300 steps, and select three everyday objects with rich interaction dynamics: backpack, plastic container, and stool.

\noindent\textbf{Metrics.}
Following InterMimic~\cite{xu2025intermimic}, we report tracking-based metrics: Success Rate (Succ.) is the percentage of reference clips successfully tracked, Duration (Time) is the average steps tracked before early termination, Robot Tracking Error ($E_r$) is the average per-link position error (cm) relative to the retargeted reference, and Object Tracking Error ($E_o$) is the average object pose/point error (cm). For failed or early-terminated clips, tracking errors are averaged over the executed frames before termination, while Succ. and Time capture whether and when the rollout fails.

\noindent\textbf{Baselines.} 
All vision-based students are distilled from the same privileged teacher (InterMimic~\cite{xu2025intermimic}) and receive the same student-side inputs, including depth, segmentation, proprioception, and the compact goal condition. Baselines differ in the visual representation module and distillation strategy. Non-world-model students replace the world model with direct visual encoders, including ResNet-18~\cite{he2016deep}, ViT~\cite{dosovitskiy2020image}, or a lightweight CNN~\cite{lecun2002gradient}, and are trained with RL-only, DAgger, or DAgger+RL. We additionally include a \emph{single-stage Dreamer} baseline~\cite{hafner2019dream,hafner2020mastering} that learns a visuomotor policy end-to-end from the same visual observations and task commands, using the same task reward and termination criteria, without privileged teacher-state inputs or teacher supervision.

\vspace{-2mm}
\subsection{Quantitative Evaluation}
\vspace{-1mm}

We evaluate distillation performance on OMOMO and BEHAVE, including stress tests with increased object mass. As shown in Table~\ref{tab:distill_smplx} and Table~\ref{tab:behave}, teacher-guided distillation achieves substantially higher success rates and longer sustained interactions than RL-only visual baselines under matched student-side inputs. On OMOMO, DreamMimic reaches \(92.2\%\) success with lower robot and object tracking errors than direct visual-encoder students. Under increased object mass, the gain is more modest but remains visible in success rate and object tracking error. Ablation results in Table~\ref{tab:distill_smplx} further indicate that action-conditioned latent consistency and predictive policy inputs both contribute to stable tracking. Although single-stage Dreamer slightly improves execution time over direct model-free RL from pixels, it remains far below teacher-guided distillation on these contact-rich humanoid tasks.

\begin{table}[t]
\centering
\resizebox{\columnwidth}{!}{
\begin{tabular}{lc*{8}{c}}
\toprule
\multirow{2}{*}{Method} & \multirow{2}{*}{Distill.}
& \multicolumn{4}{c}{SMPL-X on OMOMO}
& \multicolumn{4}{c}{SMPL-X on OMOMO (w x5)} \\
\cmidrule(lr){3-6} \cmidrule(lr){7-10}
&
& Succ.$^\uparrow$ & Time$^\uparrow$ & $E_r$$^\downarrow$ & $E_o$$^\downarrow$
& Succ.$^\uparrow$ & Time$^\uparrow$ & $E_r$$^\downarrow$ & $E_o$$^\downarrow$ \\
\midrule

InterMimic~\cite{xu2025intermimic} & Teacher
&100.0 & 190.51 & 9.4 & 6.8 
& 68.6 & 157.96 & 11.7 & 15.0 \\

\midrule
ResNet-18\cite{he2016deep} + policy & RL
& 0.0 & 18.53 & 19.8 & -
& 0.0 & 18.51 & 20.0 & - \\

Dreamer (single-stage) & RL
& 0.0 & 28.73 & 25.6 & -
& 0.0 & 28.22 & 25.9 & - \\

ResNet-18\cite{he2016deep} + policy & DAgger
& 66.7 & 163.22 & 8.9 & 10.1
& 25.5 & 105.98 & 11.3 & 16.0 \\

ResNet-18\cite{he2016deep} + policy & DAgger+RL
&72.6 & 169.49 & 7.8 & \rd 9.7
& 29.4 & 102.92 & 10.5 & 15.7 \\

Simple-CNN\cite{lecun2002gradient} + policy & DAgger+RL
& 76.5 & 173.82 & 7.4 & \rd 9.7
& 29.4 & 115.06 & 11.5 & 16.5 \\

ViT\cite{dosovitskiy2020image} + policy & DAgger+RL
& 72.6 & 166.69 & 6.4 & 10.1
& 27.5 & 107.96 & 9.5 & \rd 15.2 \\

\midrule



Without multi-step latent distill. & DAgger+RL
& 70.6 & 178.69 &7.9 & 12.8
&\nd \sl 39.2 & 117.62&9.2 &15.9 \\

RecH-only WM & DAgger+RL
& 86.3 & 177.88 &7.5 & 12.7
& 31.4 & 117.78 &9.3 &16.4 \\

RecH + object-state pred. & DAgger+RL
& 86.3 & 178.06 &6.2 & 10.0
&\nd \sl39.2 & 117.69 &8.5 & \nd \sl 15.1 \\

Auxiliary heads as losses only & DAgger+RL
& 84.3 & 182.22 &6.2 & \fs \bf 8.8
&33.3 & 116.92 &8.9 & 16.1 \\

Stochastic latent feature only & DAgger+RL
& 74.5 & 174.98 &7.7 & 13.2
&31.4 & 110.90 &8.9 & 16.1 \\

RSSM w/o action conditioning & DAgger+RL
& \nd \sl90.2 &\rd 183.08  &\rd 5.8 & \rd 9.7 
&\nd \sl39.2 & \fs \bf121.59 &\rd 8.3 &15.5  \\

RSSM w/o recurrent deter. state & DAgger+RL
&\rd88.2&\nd \sl184.12 & 6.0 &9.8
&\rd37.3 & \nd \sl120.37 &8.5 &15.8\\

Object-pose-goal-only policy & DAgger+RL
& \rd 88.2 & \nd \sl184.12 &\nd \sl5.7 & \nd \sl 9.3
&\nd \sl39.2 & 116.45 &\fs \bf 8.0 &\fs \bf 14.8 \\

\textbf{DreamMimic} & DAgger+RL
& \fs \bf 92.2 &\fs \bf 184.18 &\fs \bf 5.4 & \fs \bf 8.8
&\fs \bf 41.2 & \rd 120.29 &\nd \sl 8.1 &\fs \bf 14.8 \\

\bottomrule
\end{tabular}}
\caption{Comparison of visuomotor policies on SMPL-X in OMOMO dataset.}
\label{tab:distill_smplx}
\vspace{-3mm}
\end{table}

\begin{table}[t]
\centering
\resizebox{\columnwidth}{!}{
\begin{tabular}{lc*{8}{c}}
\toprule
\multirow{2}{*}{Method} & \multirow{2}{*}{Distill.}
& \multicolumn{4}{c}{SMPL-X on BEHAVE}
& \multicolumn{4}{c}{SMPL-X on BEHAVE (w x2)} \\
\cmidrule(lr){3-6} \cmidrule(lr){7-10}
&
& Succ.$^\uparrow$ & Time$^\uparrow$ & $E_r$$^\downarrow$ & $E_o$$^\downarrow$
& Succ.$^\uparrow$ & Time$^\uparrow$ & $E_r$$^\downarrow$ & $E_o$$^\downarrow$ \\
\midrule

InterMimic~\cite{xu2025intermimic} & Teacher
&100.0  & 304.45 & 9.7 & 10.4
& 90.1 & 303.82 & 10.3 & 12.3 \\

\midrule
ResNet-18\cite{he2016deep} + policy & RL
& 0.0 & 30.52 & 20.9 & 18.6
& 0.0 & 29.11 & 23.5 & 19.1 \\

Dreamer (single-stage) & RL
& 0.0 & 35.22 & 18.1 & 17.8
& 0.0 & 32.63 & 19.1 & \rd 18.3 \\


ResNet-18\cite{he2016deep} + policy & DAgger+RL(Naive Annealing)
& \nd \sl26.6 &\rd 130.42 &\rd 17.7 &\rd 16.2
&\nd \sl 18.2 & \rd102.33 & \rd18.8 & 18.6 \\

DreamMimic & DAgger+RL(Naive Annealing)
& \bf \fs 72.7 & \nd \sl275.21 & \nd \sl15.2 & \nd \sl14.9
&\bf \fs 63.6 & \nd \sl266.86 & \nd \sl16.8 & \nd \sl15.4 \\

\textbf{DreamMimic}  & DAgger+RL(PCG)
&\fs \bf 72.7 & \fs \bf289.53 &\fs \bf 10.2 &\fs \bf 13.3
& \fs \bf63.6 & \fs \bf268.42 & \fs \bf13.2 & \fs \bf15.1 \\

\bottomrule
\end{tabular}}
\caption{Comparison of visuomotor policies on SMPL-X in BEHAVE dataset.}
\label{tab:behave}
\vspace{-5mm}
\end{table}

\vspace{-2mm}
\subsection{Ablation Study}
\vspace{-2mm}

\paragraph{Temporal latent supervision.}
Table~\ref{tab:distill_smplx} first evaluates whether the world model should supervise only instantaneous observations or multi-step latent evolution. Removing multi-step latent distillation reduces OMOMO success from \(92.2\%\) to \(70.6\%\) and increases both robot and object tracking errors. Under increased object weight, the success gap is smaller (\(41.2\%\) vs. \(39.2\%\)), but DreamMimic still maintains better object tracking. These results suggest that action-matched imitation alone is insufficient for contact-rich rollouts; aligning future latent dynamics helps preserve temporally coherent interaction states under partial observability.

\paragraph{Auxiliary interaction prediction.}
We next isolate the supervision attached to the world model. A reconstruction-only variant (RecH-only WM) reaches \(86.3\%\) success but has larger object errors than the full model. Adding object-state prediction lowers object error from \(12.7\) to \(10.0\) cm on OMOMO and from \(16.4\) to \(15.1\) cm under increased weight. The full model further reduces object error to \(8.8\) cm on OMOMO and \(14.8\) cm under increased weight, suggesting that object-state prediction is useful but benefits from being combined with the other auxiliary signals and latent supervision.

\paragraph{Predictive features for policy conditioning.}
We further distinguish auxiliary supervision from policy conditioning. When the prediction heads are trained only as losses but their outputs are not provided to the policy, success drops to \(84.3\%\) on OMOMO and \(33.3\%\) under increased weight. The object-pose-goal-only variant achieves competitive tracking errors, especially under increased object weight, but remains below the full model in success rate. This indicates that target object pose is an important part of the goal condition, while robot trajectory cues and history-aware world-model predictions provide complementary information for acting under partial observability.

\paragraph{Recurrent dynamics conditioning.}
We ablate how temporal information enters the RSSM and the policy. Replacing the deterministic recurrent state with a current-step stochastic feature substantially reduces success to \(74.5\%\), showing that the policy benefits from a history-aware latent summary. Removing action conditioning from the RSSM causes a moderate drop from \(92.2\%\) to \(90.2\%\), whereas removing the recurrent deterministic state reduces success to \(88.2\%\). These results indicate that both action-conditioned prediction and recurrent memory matter, with the deterministic recurrent state playing the central role in maintaining stable interaction dynamics.

\paragraph{Distillation schedule.}
Following InterMimic~\cite{xu2025intermimic}, our main experiments combine DAgger-style action supervision with on-policy PPO regularization. On BEHAVE (Table~\ref{tab:behave}), we compare two teacher-guidance schedules under this shared DAgger+RL setup: iteration-based naive annealing and \emph{Performance-Conditioned Guidance} (PCG). Naive annealing is sensitive to the decay schedule because student competence under visual partial observability is difficult to predict in advance. Reducing teacher-driven rollouts too early can expose the student to out-of-distribution states, whereas delaying decay can over-anchor training to the teacher's state distribution. After tuning, naive annealing can match PCG in success rate but remains worse on tracking errors and sustained interaction duration. PCG instead adapts the teacher-environment ratio from relative teacher--student reward, which in our experiments reduces reliance on hand-tuned decay schedules while preserving supervision when the student remains under-competent.


\paragraph{Visual input modality}
We compare visual modalities for the world-model observation pipeline. On OMOMO, depth combined with segmentation achieves the strongest success rate among the tested inputs, reaching \(92.2\%\) success and 184.18 average execution steps. Segmentation-only and depth-only inputs reach \(86.3\%\) and \(88.2\%\) success, respectively, while RGB reaches \(88.2\%\) success with a slightly longer average execution time. We therefore use depth plus segmentation as the main simulated perception interface, while noting that these channels are ground-truth rendered observations rather than learned perception outputs.


%

\begin{figure}[t]
    \centering
    \includegraphics[width=0.85\columnwidth]{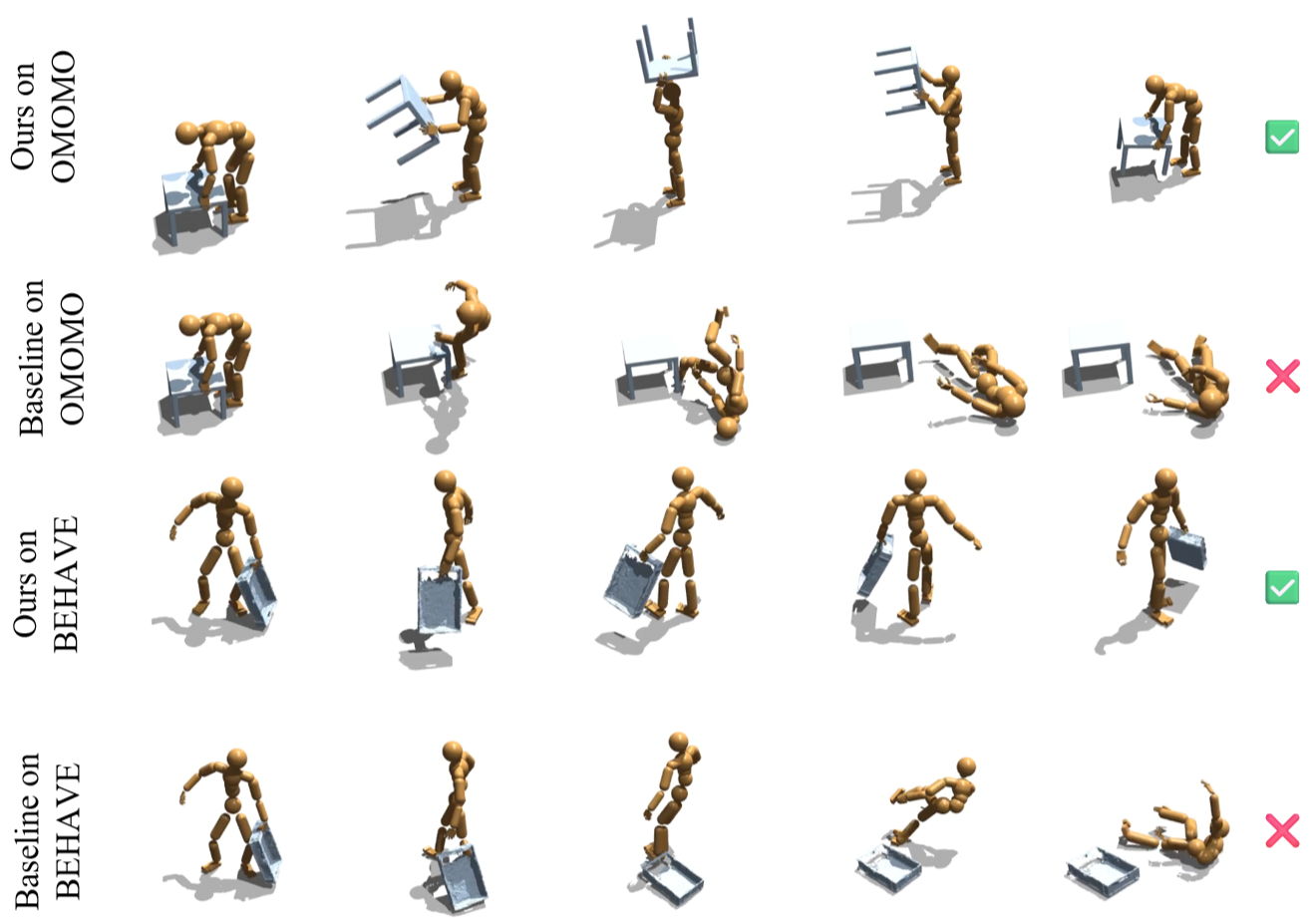}
    \caption{
    \textbf{Qualitative comparison on contact-rich loco-manipulation.}
    DreamMimic maintains stable interaction and balance over long horizons, while the baseline often fails to sustain contact and collapses.
    }
    \label{fig:qualitative_cases}
    \vspace{-4mm}
\end{figure}

\begin{figure}[t]
    \centering
    \includegraphics[width=0.9\columnwidth]{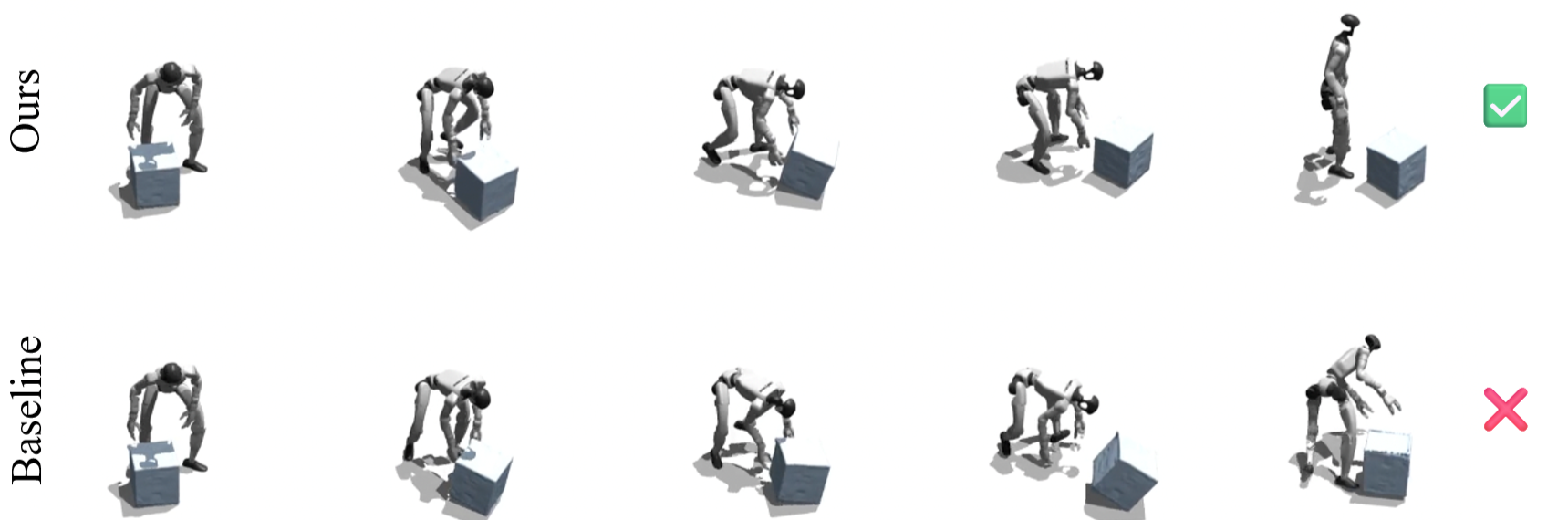}
    \caption{
    \textbf{Qualitative comparison on the Unitree G1 humanoid in Isaac Gym.}
    DreamMimic generates stable pushing behavior with consistent balance and object interaction, whereas the Dreamer baseline becomes unstable during contact and falls.
    }

    \label{fig:g1_case}
    \vspace{-4.5mm}
\end{figure}

\begin{figure}[t]
    \centering
    \includegraphics[width=0.9\columnwidth]{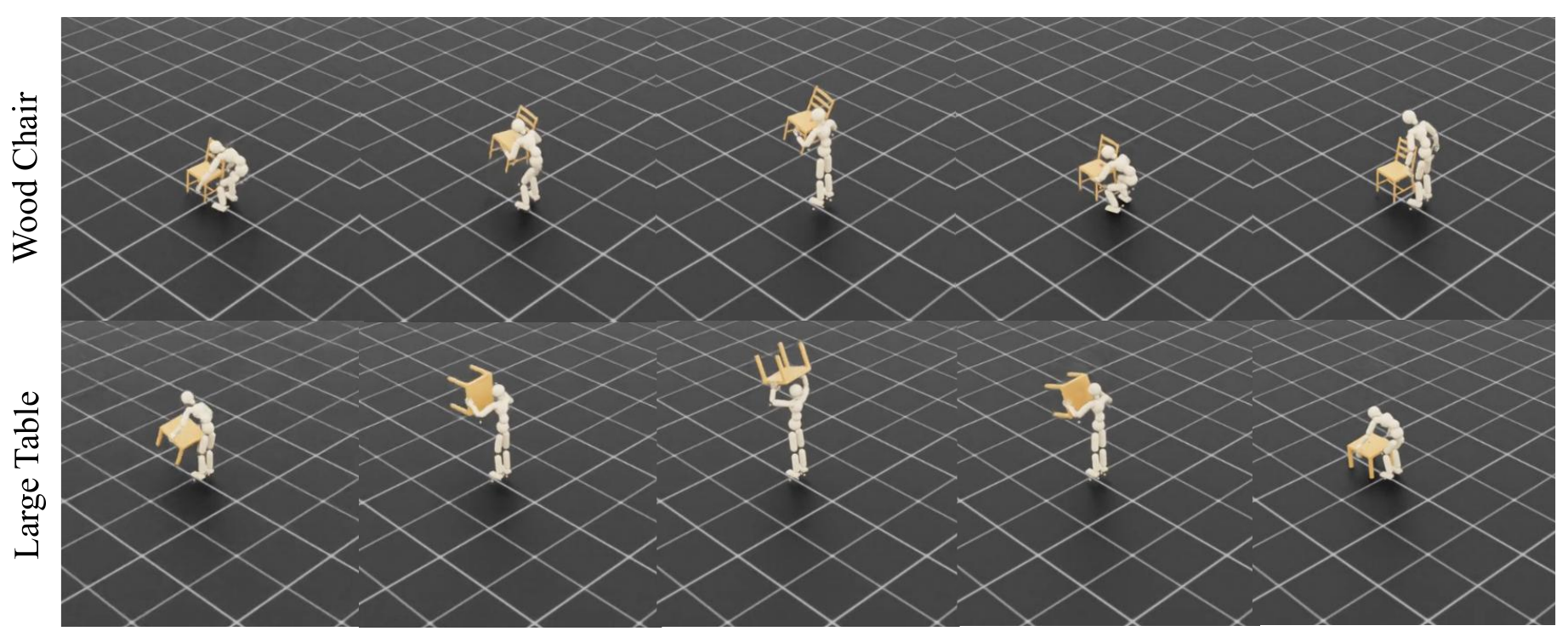}
    \caption{
    \textbf{Qualitative test in Isaac Sim.} DreamMimic performs stable whole-body interactions with a wooden chair and a large table.
    }

    \label{isaaclab}
    \vspace{-6mm}
\end{figure}

\vspace{-2mm}
\subsection{Qualitative Evaluation}
\vspace{-1mm}

\paragraph{Contact-Rich Loco-Manipulation Behaviors}
Figure~\ref{teaser} shows the vision-based student successfully completes interactions while maintaining stable whole-body contact patterns and coherent object manipulation.
As illustrated in Figure~\ref{fig:qualitative_cases}, DreamMimic executes complex object interactions that challenge the ResNet-18 policy baseline. In the OMOMO example shown in the top row of Figure~\ref{fig:qualitative_cases}, DreamMimic lifts and transports a large table while preserving balance and coordinated motion. By contrast, the baseline method depicted in the second row fails to establish stable contact and collapses. A similar trend is observed in the BEHAVE sequence, where the third row shows DreamMimic performing coordinated dragging and walking with sustained object contact. Meanwhile, the baseline in the bottom row initially grasps the container but subsequently loses contact, leading to a fall. These observations are consistent with the quantitative results, suggesting that predictive latent modeling helps maintain temporally coherent behavior in contact-rich humanoid control.

\paragraph{Cross-Embodiment Simulation} \label{sec:exp_g1}
We further evaluate DreamMimic on the Unitree G1 humanoid with 42 DoF in simulation to examine behavior under a different morphology and actuation configuration. As shown in Figure~\ref{fig:g1_case}, DreamMimic generates stable pushing behaviors with consistent balance and controlled object interaction, whereas the Dreamer baseline becomes unstable during contact and ultimately falls. This comparison provides qualitative evidence for cross-embodiment adaptation in simulation, but it should not be interpreted as hardware validation.

\paragraph{Cross-Simulator Qualitative Test} \label{sec:exp_sim2sim}
To assess cross-simulator behavior, we train policies in Isaac Gym and evaluate them in Isaac Lab under matched OMOMO sequences. As shown in Figure~\ref{isaaclab}, DreamMimic completes the selected qualitative tasks after transfer, suggesting that the learned latent dynamics capture interaction-relevant structure to some extent. This result should be interpreted as qualitative sim-to-sim evidence rather than a substitute for real-robot validation.

Additional qualitative results and full video demonstrations are available on the project website: \url{https://dreammimic.github.io/}.

\vspace{-3.5mm}
\subsection{Failure Case Analysis}
\vspace{-1mm}

Despite these improvements, several failure cases persist, especially under severe occlusion or visually ambiguous contacts where weak perceptual evidence leads to policy errors despite multi-step supervision. 
In the G1 experiments, we also observe manipulation-related breakdowns: although the teacher, trained on motions retargeted by GMR~\cite{joao2025gmr}, produces stable lower-body locomotion, upper-body hand contacts lack sufficient precision for forceful manipulation, so the policy tends to push heavy objects rather than lift them. This suggests that retargeting may preserve global motion while missing fine-grained hand--object interaction dynamics.
Finally, our evaluation is confined to simulation with ground-truth depth and segmentation, and we do not report real-robot experiments. Addressing these limitations will likely require more robust perception, tactile or local hand--object control, and recovery behaviors under occlusion.

\vspace{-2mm}
\section{Conclusion}
\vspace{-1mm}
This work presented DreamMimic, a world-model-assisted framework for stabilizing vision-based policy distillation in humanoid loco-manipulation. By learning predictive RSSM latent dynamics and applying action-conditioned multi-step latent consistency, DreamMimic addresses long-horizon drift under partial observability. We further use auxiliary supervision for interaction- and task-related quantities, together with a competence-aware guidance schedule (PCG) that adapts teacher involvement based on reward feedback. Experiments show improved tracking-based loco-manipulation performance over vision-based baselines, along with qualitative evidence under cross-embodiment and cross-simulator simulation settings.





\small
\bibliographystyle{IEEEtrans}
\bibliography{root}

\end{document}